%% file: main.tex
\documentclass[10pt,twocolumn,letterpaper]{article}

\usepackage{iccv}              
\usepackage{booktabs} 
\usepackage{makecell} 
\usepackage{multirow} 
\usepackage{diagbox}
\usepackage{pifont}


\definecolor{iccvblue}{rgb}{0.21,0.49,0.74}
\usepackage[pagebackref,breaklinks,colorlinks,allcolors=iccvblue]{hyperref}

\def\paperID{6981} 
\def\confName{ICCV}
\def\confYear{2025}

\title{
    \raisebox{-0.2cm}{\includegraphics[width=0.8cm]{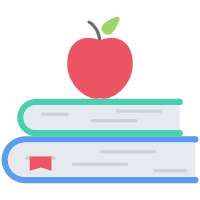}}\hspace{0.0cm}
    Knowledge Transfer from Interaction Learning 
}

\author{
Yilin Gao$^1$ \quad
Kangyi Chen$^1$ \quad
Zhongxing Peng$^1$ \quad
Hengjie Lu$^1$ \quad
Shugong Xu$^{2,}$\thanks{Corresponding author} \quad \\
$^1$Shanghai University \quad
$^2$Xi'an Jiaotong-Liverpool University \quad \\
{\tt\small $^1$\{gaoyilin, 2582712917, pzx, luhengjie\}@shu.edu.cn} \quad
{\tt\small $^2$shugong.xu@xjtlu.edu.cn}
}

\begin{document}

\maketitle
\input{sec/0_abstract}    
\input{sec/1_intro}
\input{sec/2_related}

\input{sec/3_method}
\input{sec/4_exp}
\input{sec/5_con}
\input{sec/6_ack}
{
    \small
    \bibliographystyle{ieeenat_fullname}
    \bibliography{main}
}
\input{sec/X_suppl}

\end{document}

%% file: sec/0_abstract.tex
\begin{abstract}
Current visual foundation models (VFMs) face a fundamental limitation in transferring knowledge from vision language models (VLMs): while VLMs excel at modeling cross-modal interactions through unified representation spaces, existing VFMs predominantly adopt \textit{result-oriented} paradigms that neglect the underlying interaction processes. This representational discrepancy hinders effective knowledge transfer and limits generalization across diverse vision tasks.
We propose \textbf{Learning from Interactions} (LFI), a cognitive-inspired framework that addresses this gap by explicitly modeling visual understanding as an interactive process. Our key insight is that capturing the dynamic interaction patterns encoded in pre-trained VLMs — beyond their final representations — enables more faithful and efficient knowledge transfer to VFMs. The approach centers on two technical innovations: (1) \textit{Interaction Queries}, which maintain persistent relational structures across network layers, and (2) \textit{interaction-based supervision}, derived from the cross-modal attention mechanisms of VLMs.
Comprehensive experiments demonstrate consistent improvements across multiple benchmarks: achieving $\sim$3.3\% and $+$1.6 mAP/$+$2.4 $AP^{mask}$ absolute gains on TinyImageNet classification and COCO detection/segmentation respectively, with minimal parameter overhead and faster convergence (7$\times$ speedup). The framework particularly excels in cross-domain settings, delivering $\sim$2.4\% and $\sim$9.3\% zero-shot improvements on PACS and VLCS. Human evaluations further confirm its cognitive alignment, outperforming result-oriented methods by 2.7$\times$ in semantic consistency metrics.
\end{abstract}

%% file: sec/1_intro.tex
\section{Introduction}
\label{sec:intro}

\begin{quote}
    \textit{``To teach someone how to fish is better than to just give him a fish."}
    \begin{flushright}
        -- \textit{Huainanzi} 
    \end{flushright}
\end{quote}

Recent advancements in Multi-modal Large Language Models (MLLMs) \citep{achiam2023gpt,anil2023gemini} have marked a transformative era in Artificial Intelligence (AI), enabling unprecedented capabilities in processing and integrating diverse data modalities such as text, images, sound, and video. These models have demonstrated remarkable proficiency in understanding and synthesizing human-generated content, thereby absorbing a vast spectrum of human knowledge. However, despite these achievements, a critical challenge remains: facilitating \textbf{effective knowledge transfer} between heterogeneous models while ensuring \textbf{cognitive alignment}. Current methodologies often prioritize superficial output imitation, neglecting the fundamental processes that underlie knowledge acquisition.

\begin{figure}[!tpb]
    \centering
    \includegraphics[width=0.9\linewidth]{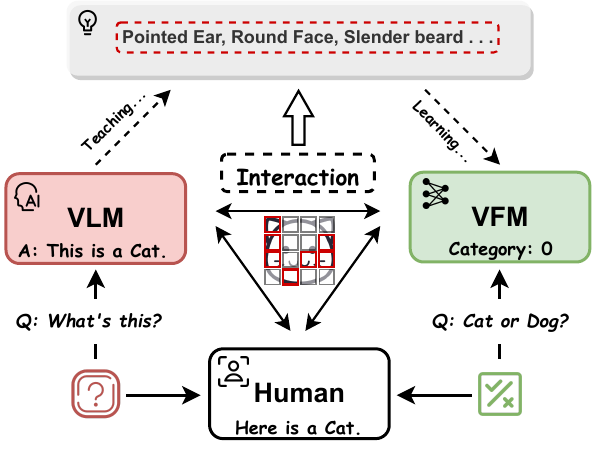}%
    \caption{
        Humans and intelligent agents, including Vision-Language Models (VLMs) and Visual Foundation Models (VFMs), comprehend the world through analogous cognitive processes, irrespective of the final form of representation. We denote this cognitive process as \textbf{\textit{interaction}}. By refining these representation-agnostic interactions, we facilitate cross-modal and cross-task knowledge transfer.
    }
    \label{img:intro}
\end{figure}

Amidst this rapid progress, a fundamental question arises: Do these models truly acquire *knowledge*? And is the cognitive process of *knowledge* consistent across different models? While MLLMs exhibit formidable capabilities, the mechanisms underlying their knowledge acquisition remain underexplored. Knowledge, being an abstract concept, is inherently difficult to define. Its representation varies across models and even among humans, manifesting through language, writing, painting, and other forms. Consequently, measuring knowledge solely based on its representational form is inherently problematic.

Although knowledge itself may not be directly measurable, the cognitive processes that facilitate its understanding can be represented. For instance, as illustrated in Figure \ref{img:intro}, both humans and models recognize a cat through a set of definitive interactions (e.g., pointed ears, round face, long whiskers). This perspective is supported by extensive research on model interpretability \citep{lake2023human,niu2024large}. We posit that these definitive interaction sets constitute an alternative representation of knowledge, and importantly, that this cognitive process is quantifiable. Following \cite{chen2023interpreting}, we term these sets \textbf{\textit{interaction}}. The essence of interactions lies in their shareability and complementarity \cite{huh2024platonic,pang2023frozen_reb,shen2023can}, implying that the cognitive processes of different agents regarding the same knowledge are analogous and can mutually enhance one another. This mirrors the pedagogical process where an experienced teacher translates knowledge into concepts and effectively imparts them to students.

Building on this understanding, we argue that although Vision Language Models (VLMs) and Visual Foundation Models (VFMs) differ in their representational approaches, VLMs can still impart their cognitive processes of knowledge to VFMs through an interaction-based perspective. This approach diverges from traditional result- or representation-oriented learning methods that rely on explicit labels. Moreover, given the extensive multi-modal knowledge (e.g., vision and language) absorbed during VLM training, VLMs can function as seasoned educators, refining the cognitive processes of VFMs. This aligns with the adage, ``teach someone to fish." To operationalize these ideas, we propose a novel knowledge transfer methodology termed \textbf{\textit{Learning from Interaction}}. By emulating the cognitive processes of VLMs for various tasks, this methodology facilitates knowledge transfer from VLMs to VFMs. The primary goal is to leverage the diverse knowledge sources of VLMs to enhance VFMs, enabling them to comprehend the natural world more holistically.

The contributions of this work are threefold:
\begin{enumerate}
    \item We empirically validate the consistency of the cognitive process of knowledge at the interaction level between VLMs and VFMs.
    \item We introduce \textbf{Interaction Learning}, a novel framework that enhances the task cognition capabilities of VFMs through additional interactive supervision from VLMs, enabling cross-task and cross-modal knowledge transfer.
    \item We demonstrate the effectiveness of our approach on downstream visual foundation tasks. Through Interaction Learning, the model exhibits significant improvements in accuracy, convergence speed, and generalization. Human evaluations further confirm that our approach aligns more closely with human cognitive processes for tasks.
\end{enumerate}

%% file: sec/2_related.tex
\section{Related Works}
\label{sec:relate}

\subsection{Foundation Visual Models}

The transformer architecture \citep{vaswani2017attention} has revolutionized computer vision, with numerous adaptations tailored to diverse tasks. The Vision Transformer (ViT) \citep{dosovitskiy2020image} pioneered the use of transformers in vision by segmenting images into patches as inputs. The Swin Transformer \citep{liu2021swin} enhanced efficiency through shifted windows, while the DEtection TRansformer (DETR) \citep{detr} extended transformers to object detection, eliminating the need for Non-Maximum Suppression (NMS) and anchor designs. Subsequent works \citep{zhu2020deformable, meng2021conditional, liu2022dab, li2022dn, zhang2022dino} addressed slow convergence by refining object queries and Hungarian matching algorithms. These advancements underscore the versatility of transformers, unifying model architectures across tasks.

\subsection{Vision Language Models}

Vision Language Models (VLMs) \citep{li2023blip,dai2023instructblip,zhu2023minigpt,liu2024visual,liu2024improved,wang2023cogvlm,chen2024internvl,li2024mini,yao2024deco,zhang2024sparsevlm} bridge visual and textual modalities, integrating diverse perceptual and representational approaches. BLIP-2 \citep{li2023blip} introduced the Q-Former to align these modalities, while LLaVA and MiniGPT-4 \citep{zhu2023minigpt,liu2024visual,liu2024improved} employed linear projection layers for alignment. Mini-Gemini \citep{li2024mini} enhanced VLMs with an additional visual encoder for high-resolution refinement. Despite their strengths, VLMs' large scale limits their practical application. DeCo \citep{yao2024deco} addressed this by compressing visual tokens at the patch level, and SparseVLM \citep{zhang2024sparsevlm} reused self-attention matrices to prune insignificant vision tokens. These works inspire us to transfer VLMs' rich cognitive understanding to Visual Foundation Models (VFMs), enhancing their comprehension of the world.

\subsection{Model Interpretability}

Model interpretability is crucial for understanding the inner workings of Deep Neural Networks (DNNs), which are often regarded as \textit{black-box} systems. Recent research has increasingly focused on this area. \citep{li2023does,kim2024does} demonstrated that DNNs can learn symbolic interactions, distilling transferable knowledge from raw data. Similarly, transformer-based vision models exhibit interactions between patches, as analyzed by \citep{chen2023interpreting,hu2024interpreting,ma2023visualizing,abnar2022quantifying,chefer2021transformer,chefer2021generic,assran2023self}. Multi-modal systems have also been explored \citep{xiao2024seeing,rasekh2024ecor,lyu2022dime,stan2024lvlm}, with LVLM-Interpret \citep{stan2024lvlm} designing interactive tools to uncover the mechanisms of large vision-language models. These efforts collectively aim to reveal the cognitive processes, or \textit{interactions}, between inputs and outputs.

\subsection{Knowledge Transfer}

Knowledge distillation, introduced by \citep{hinton2015distilling}, is a widely used method for knowledge transfer, enabling student networks to learn feature distributions or labels from teacher networks. Early works like \citep{zagoruyko2016paying} used attention maps for transfer, while \citep{chen2021distilling} leveraged multi-level teacher information. In open-vocabulary object detection, \citep{gu2021open,ma2022open,bangalath2022bridging} distilled VLM knowledge to align region-level and image-level embeddings. Recent works \citep{shu2024llava,cai2024llava} extended distillation to transfer knowledge between large-scale and small-scale VLMs. However, these methods are limited to single-modality distillation and cannot achieve cross-modal or cross-task transfer.

Building on the success of Vision-Language Model pretraining, transfer learning has become a dominant approach due to its scalability and efficiency. Techniques like prompt tuning \citep{zhou2022learning,zhou2022conditional} and visual adaptation \citep{gao2024clip,zhang2021tip} adapt pre-trained VLMs for downstream tasks. \citep{pang2023frozen} demonstrated that pre-trained LLMs can handle purely visual tasks without language reliance, while LOAT \citep{lin2024advancing} combined LLMs with historical experiences for commonsense reasoning. Inspired by these works, we propose leveraging the extensive knowledge embedded in VLMs and transferring it to VFMs to enhance their capabilities in downstream tasks.

%% file: sec/3_method.tex
\begin{figure*}[!tpb]
    \centering
    \includegraphics[width=1\linewidth]{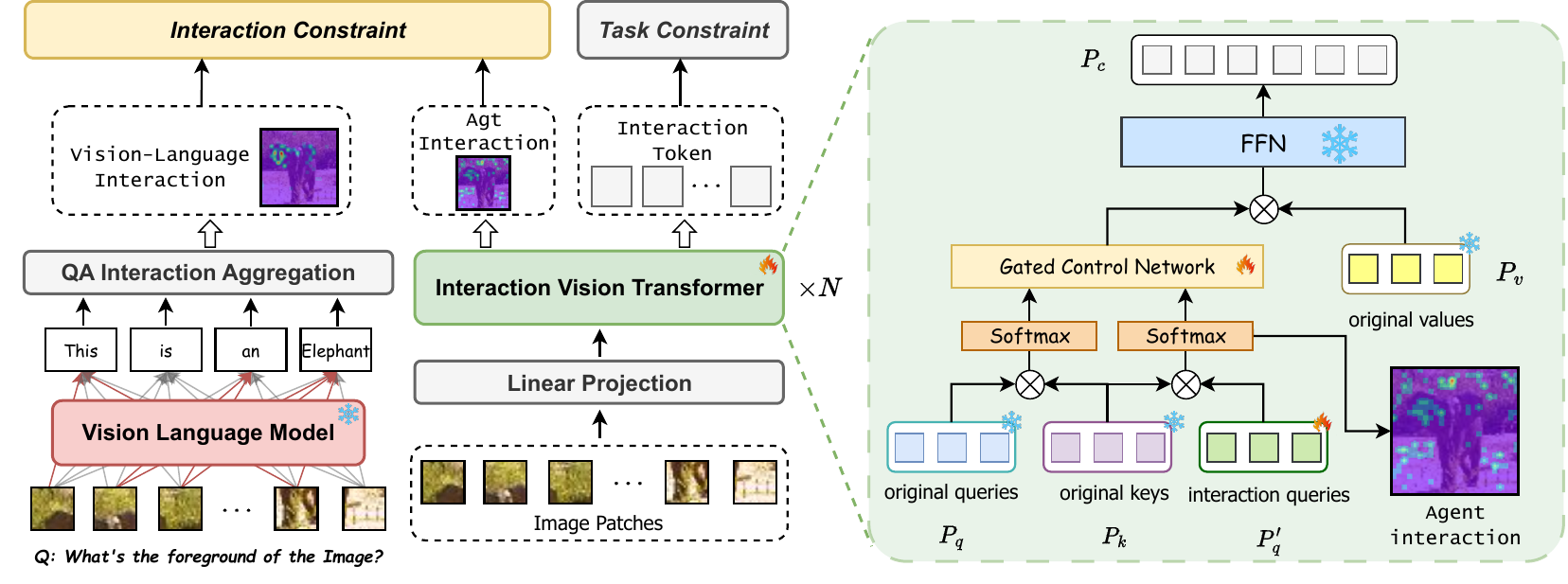}%
    \caption{
        \textbf{Pipeline of Learning from Interaction.} The process consists of several steps: (1) Extracting Vision-Language Interactions from pre-trained Vision Language Models (VLMs) using a Visual Question Answering (VQA) framework; (2) Generating additional interaction queries during query formulation in the Vision Transformer, supervised by the extracted Vision-Language Interactions; (3) Integrating both sets of interactions via a Gated Control Network; (4) Producing the final Interaction Tokens, which serve as the basis for task predictions.
    }
    \label{img:pipline}
\end{figure*}

\section{Preliminary}

\subsection{Vision Transformer}
\label{sec:vit}
The Vision Transformer (ViT) \cite{dosovitskiy2020image} processes an image as a sequence of non-overlapping patches. For an image divided into $N$ patches $\{p_1,...,p_N\}$, each patch is linearly projected into a token:
\[
\mathbf{x}_i = \text{Embedding}(p_i) + \mathbf{e}_i,
\]
where $\mathbf{e}_i$ denotes positional embeddings. These tokens are processed through stacked Transformer layers, where the multi-head self-attention mechanism computes dependencies between tokens:
\[
\text{Attention}(Q,K,V) = \text{softmax}\left(\frac{QK^\top}{\sqrt{d_k}}\right)V.
\]
This operation can be factorized into two components by identity transformation:
\begin{equation}
\text{Attention}(Q,K,V) = \underbrace{\sigma(Q,K)}_{\text{Structure}} \odot \underbrace{\phi\left(\frac{QK^\top}{\sqrt{d_k}}\right)}_{\text{Strength}} \cdot V,
\label{eq:vit_decomp}
\end{equation}
where:
- $\sigma(Q,K) \in \{0,1\}^{N \times N}$ is a binary matrix indicating token relationships (1: related, 0: unrelated),
- $\phi(\cdot)$ computes normalized attention weights,
- $\odot$ denotes element-wise multiplication.

\subsection{Definition of Interaction}
\label{sec:interaction_def}
To establish a unified framework for cross-modal knowledge transfer, we first formalize the concept of \textbf{interaction} – the fundamental cognitive process underlying both human and machine perception. Consider the human process of recognizing a cat: it emerges from synergistic relationships between features like pointed ears (A), round face (B), and long whiskers (C). These features do not act in isolation; their combinatorial relationships (e.g., ``A \textit{AND} B \textit{AND} C") collectively define the concept. Analogously, in deep neural networks, outputs are determined by structured interactions between input elements.  

Recent work \cite{li2023does} rigorously proves that the output of any DNN \( v: \mathbb{R}^n \to \mathbb{R} \) can be decomposed into logical interactions:  
\[
v(p) = \sum_{S \in \Omega_{\text{and}}} \underbrace{I_{\text{and}}(S|p)}_{\text{Conjunctive Interaction}} + \sum_{S \in \Omega_{\text{or}}} \underbrace{I_{\text{or}}(S|p)}_{\text{Disjunctive Interaction}},  
\]  
where \( I_{\text{and}} \) encodes \textbf{joint activation} of variable subsets \( S \) (e.g., ``A \textit{AND} B"), while \( I_{\text{or}} \) represents \textbf{alternative activation} (e.g., ``A \textit{OR} B"). The sets \( \Omega_{\text{and}} \) and \( \Omega_{\text{or}} \) enumerate all valid AND/OR relationships.  

\paragraph{Bridging Logical Interactions to Transformer Attention}  
In Transformer-based models, these logical interactions are explicitly manifested through attention mechanisms. To align the decomposition with architectural primitives, we re-express interactions at the token-pair level. For each attention head \( h \), the logical subsets \( S \in \Omega_{\text{and/or}} \) can be mapped to pairwise interactions between query-key pairs \( (q_i, k_j) \). Specifically:  
- \textbf{Conjunctive Interaction} (\( I_{\text{and}} \)): Implemented as the \textit{product} of binarized logic gates \( \sigma(q_i, k_j) \) and contextualized signals \( \phi(\cdot) W_o v_j \).  
- \textbf{Disjunctive Interaction} (\( I_{\text{or}} \)): Corresponds to the \textit{sum} of independent pairwise activations.  

This yields the unified formulation:  
\begin{equation}
  v(p) = \sum_{h=1}^H \sum_{i,j} \underbrace{\sigma(q_i, k_j)}_{\substack{\text{Logic Gate} \\ \text{(AND/OR)}}} \cdot \underbrace{\phi\left(\frac{q_i^\top k_j}{\sqrt{d_k}}\right) W_o v_j}_{\substack{\text{Contextualized Signal} \\ \text{(Content \& Strength)}}}.  
  \label{eq:li_interaction}
\end{equation}

Here, \( \sigma(\cdot) \) acts as a binary selector (1 for active interactions, 0 otherwise), while \( \phi(\cdot) \) modulates interaction strength. Notably, setting \( \phi(\cdot) = 1 \) simplifies the model to pure symbolic reasoning (e.g., ``A AND B"), enhancing interpretability without loss of generality.  

Critically, while \( V \) varies across tasks (e.g., classification vs. detection), the interaction structure \( C_{\text{struct}} \) and strength \( C_{\text{strength}} \) remain spatially consistent due to the shared visual grounding in natural images. This spatial coherence allows interactions to serve as universal carriers for cross-modal knowledge transfer.  

\section{Interaction Learning}
\label{sec:method}

Building on the theoretical foundation of structured interactions (Section~\ref{sec:interaction_def}), we present the Interaction Vision Transformer (I-ViT) that operationalizes cross-modal knowledge transfer through explicit interaction modeling. The overall pipeline is shown in Figure~\ref{img:pipline}.

\subsection{Interaction Vision Transformer (I-ViT)}
\label{sec:ivit}

The I-ViT enhances standard ViT through dual interaction pathways that preserve both visual and linguistic reasoning patterns. Given input tokens $\mathcal{X}$, it generates:

\begin{itemize}
    \item \textbf{Original Queries} ($P_q$): Maintain task-specific interaction structures learned from visual data
    \item \textbf{Interaction Queries} ($P_q'$): Encode cross-modal interaction patterns distilled from VLMs
\end{itemize}

\paragraph{Interaction Strength Modeling}  
Both query types interact with shared keys $P_k$ to compute interaction strength matrices:
\begin{align}
    C_{\text{VFM}} &= \phi\left(\frac{P_q P_k^\top}{\sqrt{d_k}}\right) \quad \text{(Visual Foundation Model)} \\
    C_{\text{AGT}} &= \phi\left(\frac{P_q' P_k^\top}{\sqrt{d_k}}\right) \quad \text{(VLM Guidance)}
    \label{eq:stren}
\end{align}
where $\phi(\cdot)$ denotes softmax normalization. Here, $C_{\text{VFM}}$ captures visual-centric interaction strengths, while $C_{\text{AGT}}$ incorporates linguistic-aware interaction intensities from VLMs.

\paragraph{Structural Interaction Fusion}  
The Gated Control Network (GCN) dynamically fuses interaction structures across modalities:
\begin{equation}
    [g_1, g_2] = \text{sigmoid}\left(\text{FFN}([C_{\text{AGT}}; C_{\text{VFM}}])\right),
\end{equation}
\begin{equation}
    C_F = g_1 \odot C_{\text{AGT}} + g_2 \odot C_{\text{VFM}},
\end{equation}
where $g_1,g_2 \in (0,1)$ are continuous gating weights. This implements soft structure selection:
\begin{equation}
    \text{Fused Structure} = \underbrace{g_1 \cdot \sigma_{\text{VLM}}}_{\substack{\text{Linguistic-Aware} \\ \text{Interaction Importance}}} + \underbrace{g_2 \cdot \sigma_{\text{VFM}}}_{\substack{\text{Vision-Aware} \\ \text{Interaction Importance}}},
\end{equation}
preserving the AND/OR logic through $\sigma_{\text{VLM}}, \sigma_{\text{VFM}} \in \{0,1\}$ (from Eq.~\eqref{eq:li_interaction}) while allowing adaptive importance weighting. Here:
- $\sigma_{\text{VLM}}/\sigma_{\text{VFM}}$: Binary interaction structures (0/1) from VLMs/VFMs
- $g_1/g_2$: Continuous importance scores for cross-modal alignment

This formulation separates \textit{what to interact} (binary $\sigma$) from \textit{how importantly to interact} (continuous $g$), where the former enforces logical rules and the latter enables context-aware fusion.

\paragraph{Content-Aware Interaction Generation}  
The fused interaction matrix modulates value transformations:
\begin{equation}
    P_c = \text{FFN}(C_F \odot P_v),
\end{equation}
where $P_v$ carries task-specific content information. This implements:
\begin{equation}
    \text{Interaction Tokens} = \underbrace{C_F}_{\text{Structure+Strength}} \odot \underbrace{P_v}_{\text{Content}},
\end{equation}
strictly adhering to the interaction formulation $C_{\text{struct}} \odot C_{\text{strength}} \cdot V$ from Section~\ref{sec:vit}.

\subsection{Vision-Language Interaction Extraction}
\label{sec:vlm_extract}

\paragraph{Unified Interaction Extraction Protocol}  
We establish a task-agnostic framework for extracting vision-language interactions from pre-trained VLMs, applicable to classification, detection, and segmentation. The pipeline comprises three stages:  
1. \textbf{Task-Conditioned Prompting}: Inject task semantics through language-vision prompts  
2. \textbf{Cross-Modal Activation}: Compute token-level interactions between visual and linguistic modalities  
3. \textbf{Cognitive Alignment Filtering}: Refine interactions to match human perceptual patterns  

For all tasks, the VLM processes concatenated inputs:
\begin{equation}
Q_{\text{input}} = [\mathcal{X}; \mathcal{Q}] \in \mathbb{R}^{(N+L)\times D},
\end{equation}

where $Q_{\text{input}}$ represents the concatenated sequence of image tokens $\mathcal{X}$ includes visual prompts (e.g., bounding boxes for detection) as part of the input image and question token $\mathcal{Q}$.

\paragraph{Task-Specific Formulations}  
The unified interaction extraction framework adapts to different tasks through specialized components in the response template:

\begin{itemize}
    \item \textbf{Classification}:  
    \textit{Language Prompt}: ``Classify foreground vs. background"  
    \textit{Response Structure}: \{\textit{foreground: [class], background: [...]}\}  
    \begin{equation}
    C_{\text{VLM}} = \frac{\max(0, C_{\text{fore}} - C_{\text{back}})}{\| \max(0, C_{\text{fore}} - C_{\text{back}}) \|_1}
    \end{equation}
    where $C_{\text{fore}}$ and $C_{\text{back}}$ denote interaction strengths for foreground class and background clusters respectively.
    
    \item \textbf{Dense Prediction (Detection/Segmentation)}:  
    \textit{Language Prompt}: ``Locate all \{tgt\_obj\} instances"  
    \textit{Response Structure}: \{\textit{objects: [list], background: [...]}\}  
    \begin{equation}
    C_{\text{VLM}} = \frac{\max\left(0, \sum_{k=1}^K C_{\text{obj}}^{(k)} - C_{\text{back}}\right)}{\left\| \max\left(0, \sum_{k=1}^K C_{\text{obj}}^{(k)} - C_{\text{back}}\right) \right\|_1}
    \end{equation}
    where $C_{\text{obj}}^{(k)}$ represents interaction strength for the $k$-th target object instance, aggregated across all $K$ instances.
\end{itemize}

\paragraph{Visual Prompt Integration}  
For detection/segmentation, spatial constraints \textbf{Bounding Boxes} (drawn on input images) are \textit{pre-encoded} in the input image through.

This directly shape the VLM's cross-attention patterns. As shown in Figure~\ref{fig:concept_extraction} (\textit{in Supplementary Material}), bounding boxes induce stronger activations on target regions during $C_{\text{obj}}$ computation and act as \textit{attentional anchors} that guide the VLM to:  
- Attend to object parts within prompted regions (high $C_{\text{obj}}$ values)  
- Suppress irrelevant background (near-zero $C_{\text{obj}}$)  .

\paragraph{Cognitive Filtering Mechanism}  
The operator $\frac{\max(0, \cdot)}{\|\cdot\|_1}$ implements two biological principles:  
1. \textbf{Non-Negative Competition}: Neural activations are constrained to non-negative values to align with the stability conditions of \textbf{softmax} in Equation~(\ref{eq:stren}), preventing instability.  
2. \textbf{Contrast Enhancement}: Amplify foreground-background differentiation through selective signal amplification.

\subsection{Optimization Objective}
\label{sec:optimization}

The learning objective combines task performance and cognitive alignment through a dual-loss framework:
\begin{align}
    \mathcal{L}_{\text{total}} &= \mathcal{L}_{\text{task}} + \mathcal{L}_{\text{align}}, \\
    \mathcal{L}_{\text{align}} &= D_{\text{KL}}(C_{\text{AGT}} \parallel C_{\text{VLM}}),
\end{align}
where $\mathcal{L}_{\text{task}}$ is task-specific, and $\mathcal{L}_{\text{align}}$ measures the KL-divergence between VLM-guided interactions ($C_{\text{AGT}}$) and visual foundation model's intrinsic patterns ($C_{\text{VFM}}$). The Gated Control Network dynamically adjusts the contribution of these interactions, eliminating the need for manual hyperparameter tuning.

%% file: sec/4_exp.tex
\section{Experiments}
\subsection{Experimental Setup}
\subsubsection{Datasets and Metrics.} 
Evaluations cover \textbf{TinyImageNet} (classification), \textbf{COCO}/\textbf{VOC} (detection/segmentation), and \textbf{PACS}/\textbf{VLCS} (zero-shot domain generalization). Metrics include \textit{Top-1} accuracy (classification), mAP (detection), $AP_{mask}$ (segmentation), and cross-domain variance (zero-shot).

\subsubsection{Implementation.} 
We adopt LLaVa-1.5 \cite{liu2024visual} as VLM and ViT \cite{dosovitskiy2020image} as classification model. For detection/segmentation, DETR\cite{detr} is used for COCO benchmark while Conditional-DETR\cite{conditional_detr} is used for few-shot evaluations. Training uses 4$\times$A6000 GPUs with 50 epochs (25 for segmentation), \textbf{freezing original transformer layers} and only updating the GCN, interaction queries, and task heads. Other hyperparameters follow ViT/DETR defaults.

\subsection{Experimental Results}
\subsubsection{Comparison with baseline methods.}
\begin{table}[!htpb]
\centering
\caption{Comparison of I-ViT and ViT on TinyImageNet.}
\label{tab:model_comparison}
\begin{tabular}{c|ccc}
\toprule
\textbf{Model} & \textbf{Top-1 Acc. (\%)} & \textbf{\#params} & \textbf{FLOPs} \\
\midrule
ViT-S/16 \citep{dosovitskiy2020image} & 79.94 & 22M & 15.5B \\
I-ViT-S/16 & \textbf{81.52 (+1.6)} & 24M & 18.1B \\
\midrule
ViT-B/16 & 80.24 & 86M & 55.5B \\
I-ViT-B/16 & \textbf{82.14 (+1.9)} & 93M & 62.7B \\
\midrule
ViT-L/16 & 87.83 & 304M & 191.2B \\
I-ViT-L/16 & \textbf{91.08 (+3.3)} & 329M & 213.9B \\
\bottomrule
\end{tabular}
\end{table}

\begin{figure}[!htpb]
\centering
\includegraphics[width=0.95\linewidth]{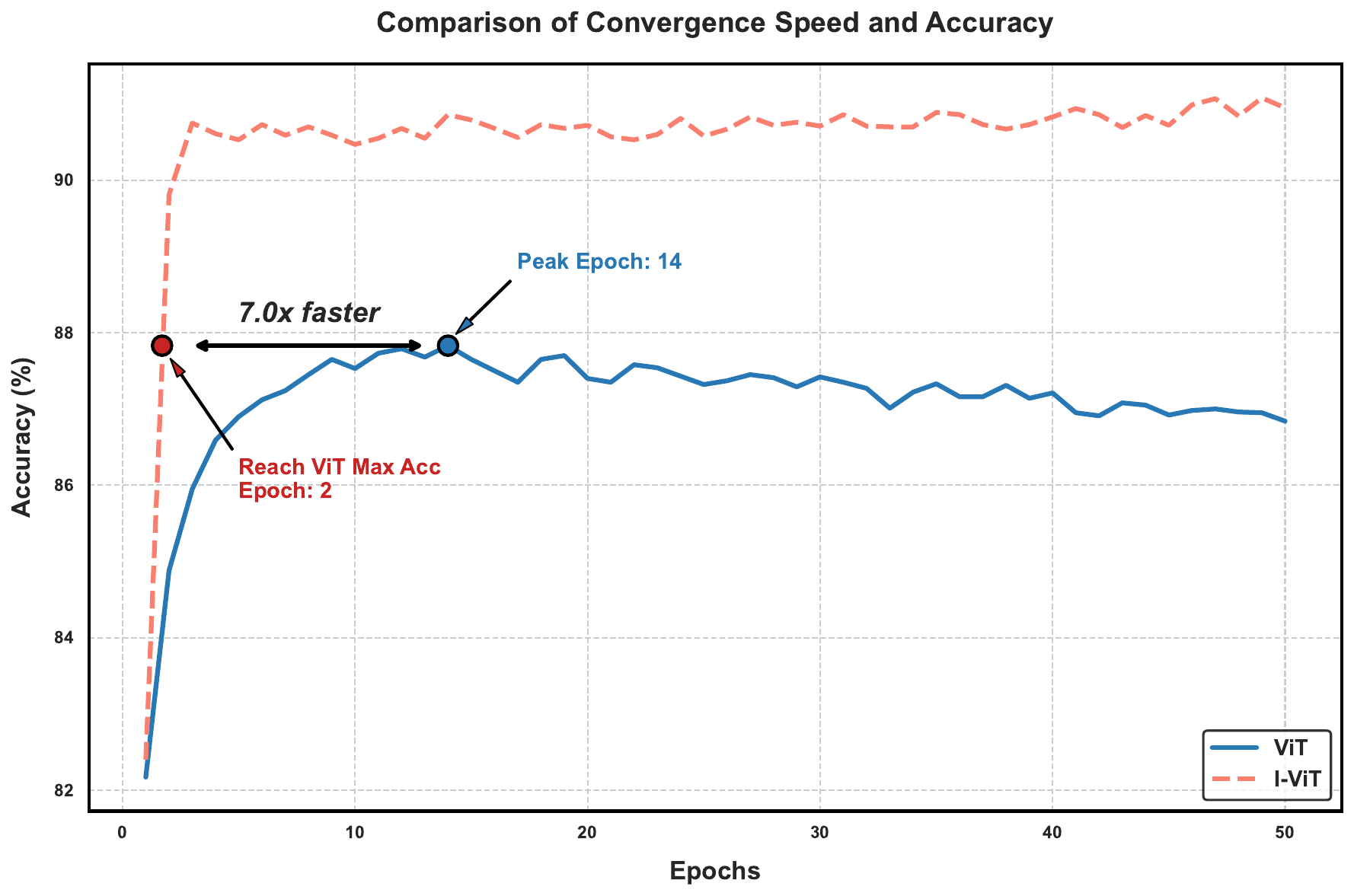}
\caption{
\textbf{Comparison of Convergence Speed.} I-ViT-L/16 exhibits faster convergence and achieves higher accuracy compared to the traditional Vision Transformer (ViT-L/16), further highlighting the efficiency of learning from interactions.}
\label{img:cov_spd}
\end{figure}

Given the limited research on knowledge transfer from VLMs to VFMs, we compared Vision Transformers (ViT) and Interaction Vision Transformers (I-ViT) on fundamental visual tasks. As shown in Table \ref{tab:model_comparison}, I-ViT significantly outperforms ViT with a reasonable increase in parameters. We also hypothesized that Interaction Learning, which focuses on teaching cognitive processes rather than results, would lead to faster convergence and better performance with fewer training samples. Figure \ref{img:cov_spd} demonstrates that I-ViT converges approximately \textbf{\textit{7x}} faster than result-oriented learning.


We further evaluated our method on reduced-scale datasets (1\%, 10\%, and 50\% of the training data). Table \ref{tab:accuracy_comparison} shows that Interaction Learning consistently achieves substantial performance gains, even with limited data. Training on 100\% of the data from scratch also confirms the indispensability of $ C_{VFM} $.

\begin{table}[htbp]
    \centering
    \caption{Comparison of I-ViT-L/16 and ViT-L/16 at Different Training Data Ratios on TinyImagenet.}
    \label{tab:accuracy_comparison}
    \begin{tabular}{>{\centering\arraybackslash}m{1.2cm} *{4}{>{\centering\arraybackslash}m{1.2cm}}}
        \toprule
        \multirow{2}{*}{\textbf{Model}} & \multicolumn{4}{c}{\textbf{Training Data Ratio (\%)}} \\
        \cmidrule{2-5}
        & 100\% & 50\% & 10\% & 1\% \\
        \midrule
        ViT & 32.67 & 27.04 & 12.64 & 2.86 \\
        I-ViT & \textbf{34.12} & \textbf{29.04} & \textbf{15.60} & \textbf{3.07} \\
        \bottomrule
    \end{tabular}
\end{table}
\subsubsection{Ablations on Interaction Vision Transformer.}

The I-ViT introduces additional queries and networks, increasing parameters and computational overhead. To demonstrate that performance improvements stem from effective learning rather than increased parameters, we conducted ablation studies, as detailed in Table \ref{tab:ablation_study}.

\begin{table}[htbp]
    \centering
    \caption{Ablations on Interaction Queries (IQ), Interaction Constraint (IC), and Gated Control (GC).}
    \label{tab:ablation_study}
    \setlength{\tabcolsep}{4pt} 
    \begin{tabular}{c|ccc|c}
        \toprule
        \textbf{Model} & \makecell[c]{\textbf{Interaction} \\ \textbf{Query}} & \makecell[c]{\textbf{Interaction} \\ \textbf{Constraint}} &  \makecell[c]{\textbf{Gated} \\ \textbf{Control}} & \textbf{Acc.} \\
        \hline
        ViT-L/16 & & & & 87.8 \\
        \multirow{4}{*}{I-ViT-L/16} & \checkmark & & \checkmark &89.6 \\
        & & \checkmark & & 80.7 \\
        & \checkmark & \checkmark & & 90.3 \\
        & \checkmark & \checkmark &\checkmark & \textbf{91.1} \\
        \bottomrule
    \end{tabular}
\end{table}


Specifically, the \textit{second row} of Table \ref{tab:ablation_study} shows the effect of removing the supervision of $C_{VLM}$ in I-ViT and introducing an extra set of learnable queries to expand the network's scale. It is observed that this additional set of learnable queries indeed enhances performance. 
However, the lack of additional supervision and the parallel operation of Interaction Queries and Original Queries mean that the network does not learn distinct and complementary information during backpropagation.


\begin{figure}[!htpb]
\centering
\includegraphics[width=0.95\linewidth]{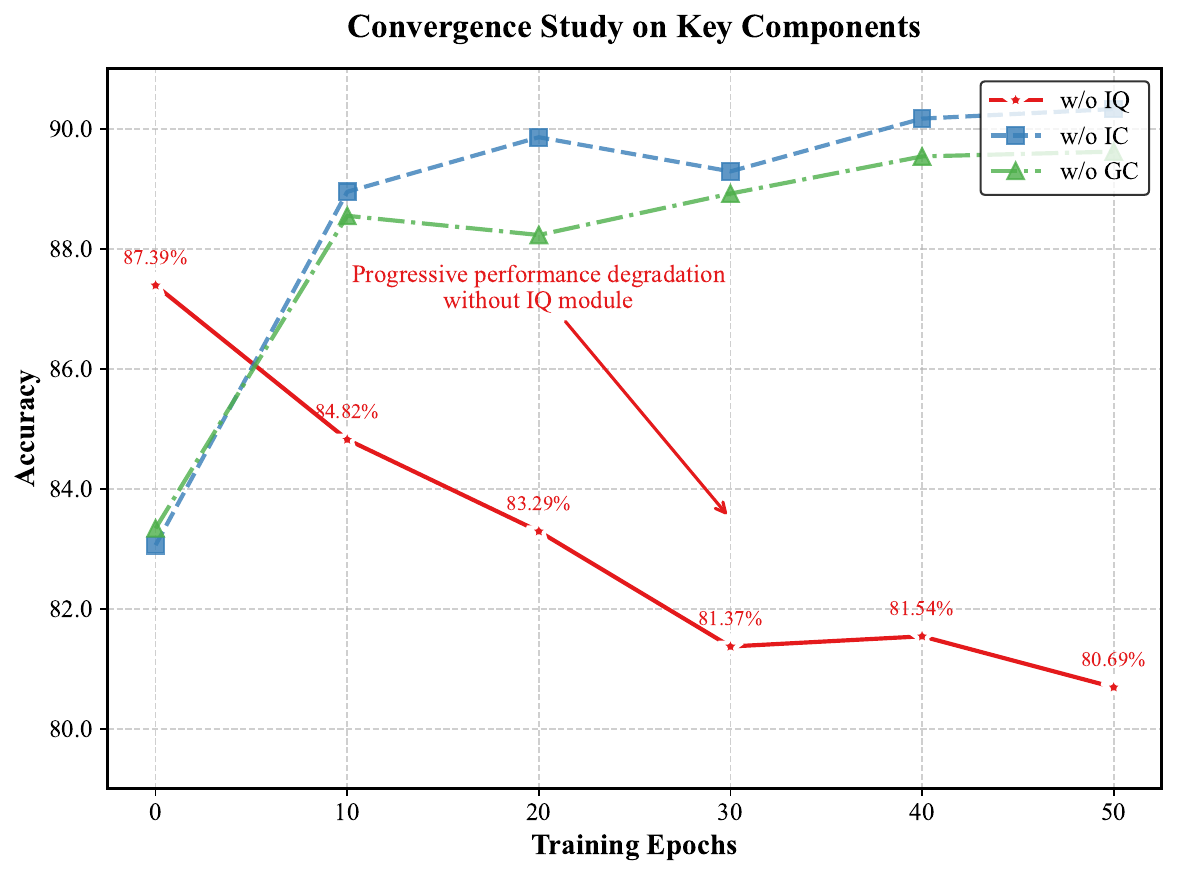}%
\caption{ 
\textbf{Component Ablation Analysis} reveals distinct failure patterns. While removing \textsc{Interaction Constraint} (w/o IC, \textcolor{blue}{dashed blue}) or \textsc{Gated Control} (w/o GC, \textcolor{green}{dash-dot green}) causes temporary accuracy drops, both variants recover to 90.3\% and 89.6\% through parameter adaptation. The \textsc{IQ}-ablated model (w/o IQ, \textcolor{red}{solid red}) shows deceptive initial competence comparable to standard ViT, benefiting from shared visual feature learning objectives. Progressive divergence between $C_{\text{VFM}}$'s task-specific representations and $C_{\text{VLM}}$'s semantic structures leads to \textbf{irreversible degradation} (6\% relative drop to 80.7\%), demonstrating \textsc{IQ}'s indispensable role in cross-modal interaction reconciliation.
}
\label{img:cov_std}
\end{figure}

Furthermore, we explored directly supervising the Original Queries with Vision-Language Interactions, as shown in the \textit{third row}. 
The results indicate not only a failure to improve performance but also a degradation, further supporting our hypothesis that interactions derived by different models for the same task, due to different training sources, are complementary rather than directly interchangeable.

As shown in Fig.~\ref{img:cov_std}, removing \textsc{Interaction Constraint} (IC) or \textsc{Gated Control} (GC) causes only transient accuracy fluctuations (90.3\% and 89.6\% final accuracy respectively), confirming their non-essential role in convergence. The critical exception is \textsc{IQ} ablation -- despite initial ViT-level performance (87.4\%), it suffers catastrophic decline to 80.7\% accuracy as modality conflicts between $C_{\text{VFM}}$ and $C_{\text{VLM}}$ escalate. This proves \textsc{IQ}'s unique necessity in sustaining stable multimodal integration.

The \textit{fourth row} validates the necessity of the Gated Control Network (GCN). By adaptively adjusting the weights of the two interactions, the model can synthesize more reliable interactions, thereby enhancing its cognitive capabilities.

\subsubsection{Visualization on Interaction}
\label{sec:vis_concept}
\begin{figure}[!htpb]
\centering
\includegraphics[width=0.95\linewidth]{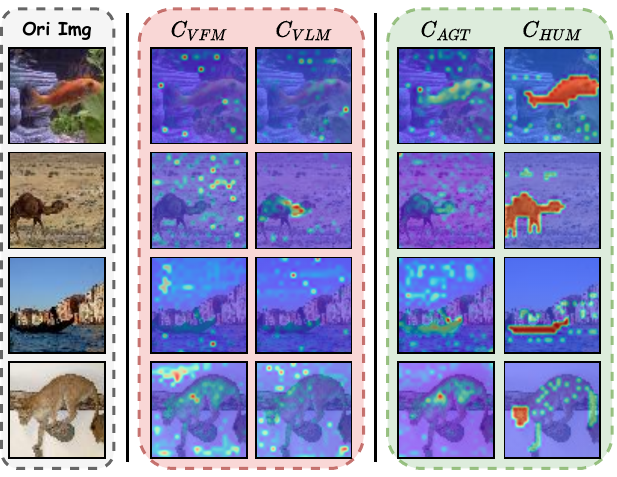}%
\caption{
\textbf{Interaction Visualization Analysis.} \( C_{\text{VFM}} \) predominantly emphasizes background information, whereas \( C_{\text{VLM}} \) concentrates more on the instances themselves. \( C_{\text{AGT}} \) synthesizes the cognitive processes and capabilities of both \( C_{\text{VFM}} \) and \( C_{\text{VLM}} \), enabling object recognition through a balance of minimal background context and a focus on the instances, thereby aligning more closely with human cognitive patterns.}

\label{img:vis_concept}
\end{figure}

In this section, we compare  \( C_{\text{VFM}} \), \( C_{\text{VLM}} \), and \( C_{\text{AGT}} \), as shown in Figure \ref{img:vis_concept}.

It is observed that $C_{\text{VFM}}$ tends to focus more on background information compared to $C_{\text{AGT}}$ and $C_{\text{VLM}}$. This bias is likely due to backgrounds occupying a larger area in the training images, leading the model to learn through background cues with minimal instance information.
Conversely, $C_{\text{VLM}}$, which benefits from extensive visual-linguistic training data, exhibits a nuanced understanding of interactions, focusing more on instances rather than environmental context.

Despite being trained on the same data as $C_{\text{VFM}}$, $C_{\text{AGT}}$ integrates the cognitive processes and capabilities of both $C_{\text{VFM}}$ and $C_{\text{VLM}}$, recognizing objects through a combination of minimal background information and a greater focus on instances. We also compared $C_{\text{AGT}}$ with the human cognitive process $C_{\text{HUM}}$, and as expected, $C_{\text{AGT}}$ aligns more closely with human cognition.
\begin{figure}[!htpb]
\centering
\includegraphics[width=0.95\linewidth]{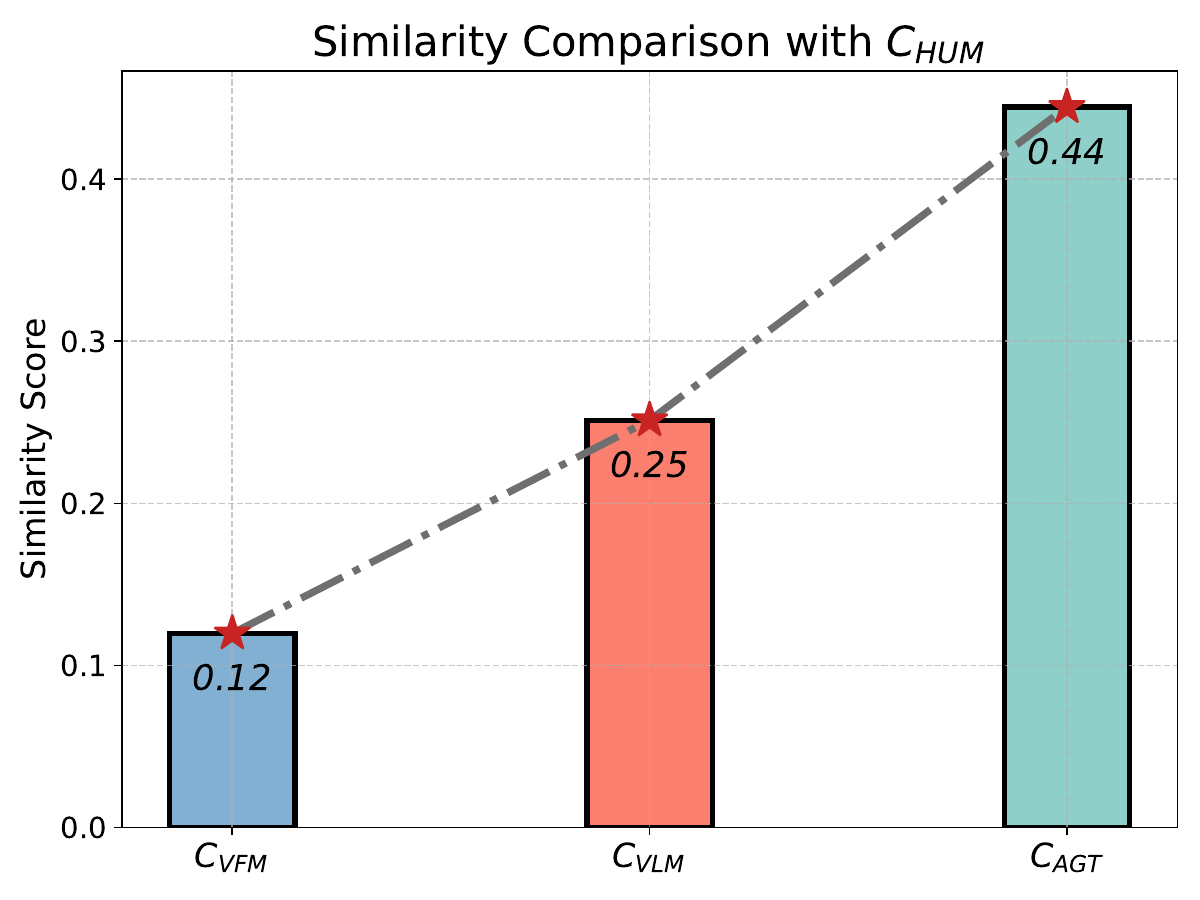}%
\caption{
\textbf{Comparison with Human Cognition ($ C_{\text{HUM}} $).} $ C_{\text{VLM}} $ is closer to human cognition than $ C_{\text{VFM}} $, but $ C_{\text{AGT}} $ achieves the highest score, effectively simulating human cognitive processes.}
\label{img:comp_score}
\end{figure}

Furthermore, we quantitatively compared the proximity of $C_{\text{VFM}}$, $C_{\text{VLM}}$, and $C_{\text{AGT}}$ to human cognitive process using \textbf{\textit{Cosine Similarity}}, as depicted in Figure \ref{img:comp_score}. Consistent with the qualitative analysis, $C_{\text{VLM}}$ is closer to the human cognitive approach than the result-oriented $C_{\text{VFM}}$. However, $C_{\text{AGT}}$ achieved a significantly higher score, effectively simulating the human cognitive process. Additional experimental settings and details are provided in Appendix \ref{sec:hum_eval}.

\subsubsection{Experiments on Cross-Domain Datasets}
To validate the generalizability of Interaction Learning, we conducted zero-shot experiments on the cross-domain dataset \textbf{PACS}. As shown in Table \ref{tab:PACS_comparison}, I-ViT outperforms ViT across nearly all domains, reinforcing the efficiency of Interaction Learning.

\begin{table}[htpb]
    \caption{\textbf{Comparison of ViT\cite{dosovitskiy2020image} and I-ViT on the PACS Dataset.} I-ViT demonstrates excellent domain generalization capabilities, validating that learning from interactions is a more efficient and robust learning approach.}
    \label{tab:PACS_comparison}
    \centering
    \begin{tabular}{c|cccc}
        \toprule
        \textbf{Model} & \makecell{\textbf{Art} \\ \textbf{Painting}} & \textbf{Cartoon} & \textbf{Photo} & \textbf{Sketch} \\
        \midrule
        ViT-L/16        & 45.28  & 29.89  & 49.86  & \textbf{15.49} \\
        I-ViT-L/16      & \textbf{51.89} & \textbf{31.90} & \textbf{51.22} & 15.18 \\
        \bottomrule
    \end{tabular}
\end{table}

On the \textbf{Art Painting} subset, which includes human-created subcategories, I-ViT achieved a $\sim 6.6\%$ improvement. This subset represents human interpretations of natural objects, with an abstraction level between \textbf{Photo} and \textbf{Sketch}. However, I-ViT experienced slight performance degradation in the \textbf{Sketch} category, likely due to the simplistic content of sketches, which makes consistent interactions challenging.

\begin{table}[htpb]
\caption{\textbf{Comparison of VLM, ViT and I-ViT on the VLCS Dataset.} I-ViT's behavioral trends align with LLaVa, demonstrating effective knowledge transfer through interaction mechanisms.}
    \label{tab:VLCS_comparison}
    \centering
    \begin{tabular}{c|cccc}
        \toprule
        \textbf{Model} & \textbf{PASC.} & \textbf{CALT.} & \textbf{LABE.} & \textbf{SUN} \\
        \midrule
        LLaVa-1.5    & 56.40 & 62.89 & \textbf{52.58} & 32.71 \\
        ViT-L/16     & 60.42 & 58.59 & 30.49 & 31.86 \\
        I-ViT-L/16   & \textbf{63.98} & \textbf{68.75} & 49.35 & \textbf{36.61} \\
        \bottomrule
    \end{tabular}
\end{table}

In VLCS (Table~\ref{tab:VLCS_comparison}), we further compared the accuracy of LLaVa. It can be found that while I-ViT surpasses ViT,  and its behavioral trends align with LLaVa, proving the effective transfer between interactions.

\subsection{Extension on Dense Prediction}
\subsubsection{Main Results on COCO}
Our experiments on COCO 2017 reveal consistent performance gains across both detection and segmentation tasks. As shown in Table~\ref{tab:coco_results}, the integration of vision-language interactions ($C_{\text{VLM}}$) elevates detection mAP from 42.0 to 43.6 (+1.6 relative improvement). Notably, the segmentation task benefits more substantially in mask-level accuracy (+2.4 AP$^{\text{mask}}$), suggesting that cross-modal cues particularly enhance boundary-sensitive predictions.

\begin{table}[htbp]
\centering
\caption{COCO 2017 Benchmark Results (val set)}
\label{tab:coco_results}
\begin{tabular}{c|c|cccc}
\toprule
\textbf{Task} & \textbf{$C_{\text{VLM}}$} & \textbf{mAP} & \textbf{AP$^{\text{mask}}$} & \textbf{\#param} & \textbf{FLOPs} \\ 
\midrule
\multirow{2}{*}{Det} 
& \checkmark & \textbf{43.6} & - & 41.9M & 55.65B \\
& \ding{55} & 42.0 & - & 41.5M & 55.62B \\ 
\hline
\multirow{2}{*}{Seg}
& \checkmark & - & \textbf{33.5} & 43.3M & 162.6B \\
& \ding{55} & - & 31.1 & 42.9M & 162.4B \\ 
\bottomrule
\end{tabular}
\end{table}

This cross-task success comes with minimal computational overhead - the added parameters less than $\sim$1\% of base models (41.9M vs 41.5M for detection). The FLOPs analysis confirms our design's efficiency, showing less than $\sim$0.6\% increase. These findings collectively validate that vision-language interactions provide generalized benefits beyond specific task formulations.

\subsubsection{Data-Efficient Learning on VOC}
The VOC experiments in Table~\ref{tab:voc_results} demonstrate our method's effectiveness in data-scarce scenarios (Condition-DETR\cite{conditional_detr}; Train from scratch). 
\begin{table}[htbp]
\centering
\caption{Few-Shot Detection Analysis on VOC}
\label{tab:voc_results}
\begin{tabular}{@{}c|cccc@{}}
\toprule
\textbf{Model} & \textbf{mAP} & \textbf{AP$_S$ }& \textbf{AP$_M$} & \textbf{AP$_L$}  \\ 
\midrule
w/ $C_{\text{VLM}}$ & \textbf{40.9} & 4.9 & \textbf{24.2} & \textbf{53.5} \\  
w/o $C_{\text{VLM}}$ & 38.6 & \textbf{5.0} & 22.4 & 51.2 \\
\bottomrule
\end{tabular}
\end{table}

With only 10\% of COCO's training data, the $C_{\text{VLM}}$ integration achieves 40.9 mAP, outperforming the baseline by 2.3 points. This significant gap (+6.0\% relative) highlights the method's ability to compensate for limited annotations through linguistic knowledge transfer.
The $C_{\text{VLM}}$ integration primarily enhances medium and large object detection ($\text{AP}_M$: +1.8, $\text{AP}_L$: +2.3), as the query attention mechanism naturally prioritizes salient regions where these objects dominate. 
Small object performance remains comparable ($\text{AP}_S$: -0.1), 
indicating preserved spatial sensitivity in the visual foundation model. This phenomenon aligns with the inherent bias of vision-language interactions toward capturing semantically prominent patterns in images.





%% file: sec/5_con.tex
\section{Conclusions and Future Work}

Multi-modal language models (MLLMs) have achieved remarkable progress in various visual tasks, largely due to the availability of large-scale image-text datasets. However, the inherent discrepancy between input and output modalities hinders the effective transfer of their natural scene understanding capabilities to downstream Visual Foundation Models (VFMs). To address this challenge, we propose \textbf{Learning from Interactions}, a novel paradigm that emphasizes the alignment of cognitive processes in knowledge transfer between Vision Language Models (VLMs) and Visual Foundation Models (VFMs). By introducing additional interaction-based supervision from VLMs to VFMs, we enable cross-modal and cross-task knowledge transfer. Our experiments demonstrate that interaction-based learning outperforms traditional result-oriented approaches in terms of accuracy, convergence speed, and generalization, while also aligning more closely with human cognitive processes.

\noindent \textbf{Future Work:}
Building upon the current framework, we focus on knowledge transfer from VLMs to VFMs. In future, we aim to explore bi-directional knowledge transfer, including from VFMs to VLMs and VFMs to VFMs, to uncover a more fundamental knowledge system and the underlying knowledge modeling process within models. Additionally, we plan to extend this approach to other foundational tasks, such as video understanding, to fully leverage the world understanding capabilities of MLLMs. This will further bridge the gap between human-like cognitive processes and machine learning models, paving the way for more robust and interpretable AI systems.

%% file: sec/6_ack.tex
\section{Acknowledgment}
This work was supported in part by the 6G Science and Technology Innovation and Future Industry Cultivation Special Project of Shanghai Municipal Science and Technology Commission under Grant 24DP1501001, in part by  the National High Quality Program under Grant TC220H07D.

%% file: sec/X_suppl.tex
\clearpage
\setcounter{page}{1}
\maketitlesupplementary

\section{Vision-Language Interaction Extraction} \label{sec:vlm-extraction}
In this section, we provide a detailed description of the Vision-Language Interaction Extraction process, as outlined in Section \ref{sec:vlm_extract}. 

\begin{figure}[!htpb]
\centering
\includegraphics[width=0.9\linewidth]{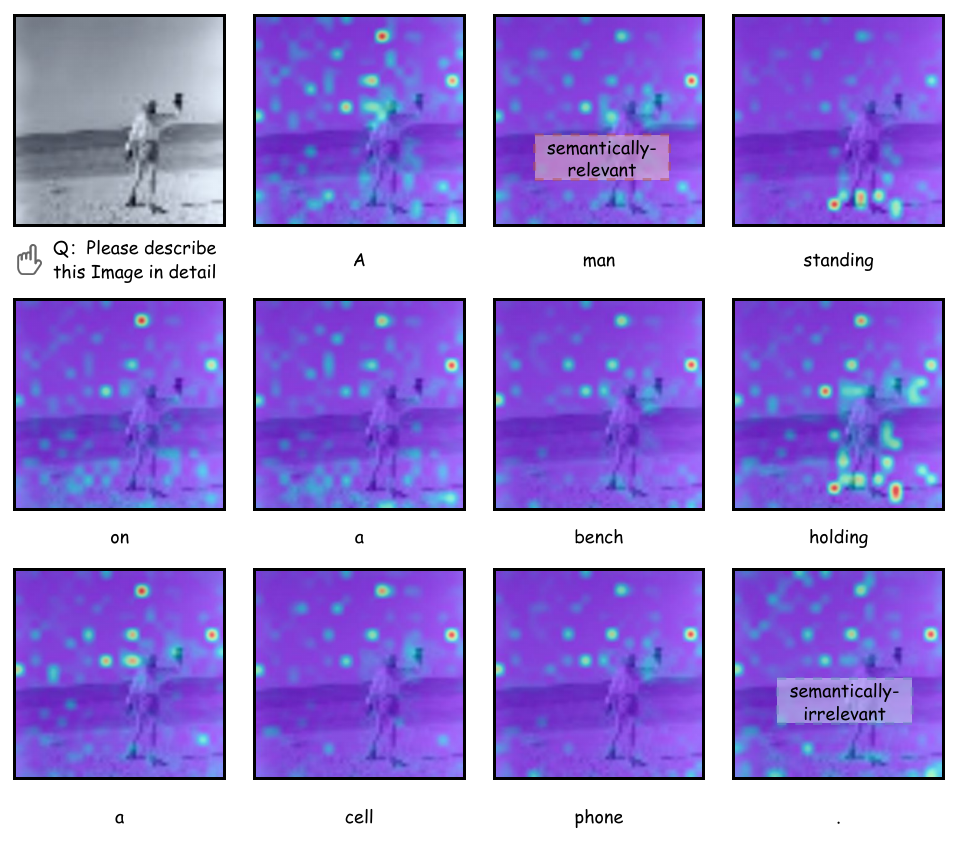}
\caption{
\textbf{Visual-Semantic Correlation in VLM Responses.} The outputs from Vision-Language Models (VLMs) are strongly associated with visual semantics. However, certain visual concepts generate high responses to semantically-irrelevant texts, such as punctuation marks, making it challenging to extract visual concepts directly from complex answers.}
\label{img:img2word}
\end{figure}

Given our use of Visual Question Answering (VQA) for interaction extraction, we first confirmed that visual concepts can effectively focus on relevant text during the VQA process, as demonstrated in Figure~\ref{img:img2word}. While VLMs exhibit strong visual-semantic correlations, we observed that visual concepts also respond to semantically-irrelevant texts, such as punctuation marks, complicating the direct extraction of visual concepts from complex answers. Therefore, developing a robust pipeline for interaction extraction via VQA is essential. We compared the concepts extracted using various prompt settings:

\begin{enumerate}
    \item \textbf{Prompt:} ``If you are doing an image classification task, what is the foreground and what is the background? The answer format is as follows: \{`foreground':\{\}, `background':\{\}\}. Please choose the foreground word from the list below:". We supplied a class vocabulary for LLaVA to select from, based on the categories in TinyImageNet. We subtracted the background interaction $ C_{\text{back}} $ from the foreground interaction $ C_{\text{fore}} $ to align the concepts with human cognitive processes, and set the minimum value to 0 and normalized the interaction scores to a range of 0 to 1, as described in Section \ref{sec:interaction_def}.
    \item \textbf{Prompt:} ``If you are doing an image classification task, What is the object in the picture? Answer the question using a single word or phrase." We directly take the interaction of the answer as our final interaction $C_{VLM}$. 
    \item  \textbf{Prompt:} ``Please describe the object in the picture in detail." Since there are several works in the sentence, we take the average of concepts as our final interaction $C_{VLM}$. 
\end{enumerate}

\begin{figure}[!htpb]
\centering
\includegraphics[width=0.9\linewidth]{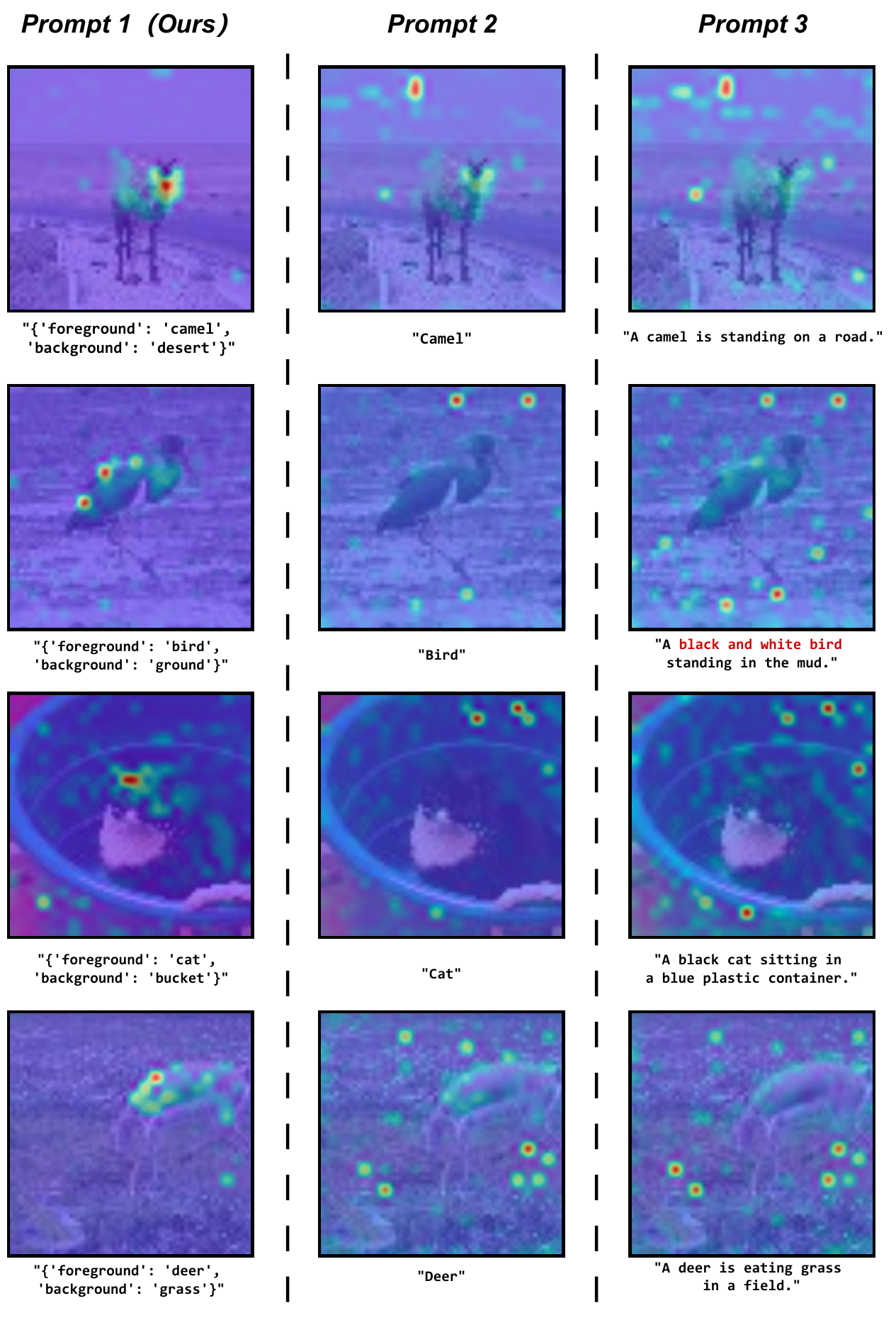}
\caption{
\textbf{Impact of Prompts on Concept Extraction.} Prompt 1 effectively focuses on the instance itself. However, Prompt 2, which instructs VLMs to output target content as single words, still leads to concepts emphasizing the background rather than the instance. Meanwhile, the detailed description approach of Prompt 3 results in concepts being distributed across the entire image and occasionally leads to incorrect responses \textcolor[RGB]{204,0,0}{\textit{(highlighted in red)}}.}

\label{fig:prompt_comparison}
\end{figure}

The results, illustrated in Figure \ref{fig:prompt_comparison}, demonstrate that different prompts elicit distinct concepts. Prompt 1 successfully emphasizes the object itself, while Prompt 2 exhibits the same issue as existing Visual Feature Models (VFMs), where concepts tend to focus on the background rather than the object. In contrast, Prompt 3's detailed description approach disperses concepts across the image, sometimes leading to inaccurate responses.

\begin{table}[!htpb]
\centering
\caption{Comparison of different Prompts}
    \begin{tabular}{ccc}
    \toprule
    Model & Prompt & Top-1 Acc.(\%) \\
    \midrule
    ViT-S/16  & /               & 79.94          \\
    I-ViT-S/16 & Prompt 3       & 80.59          \\
    I-ViT-S/16 & Prompt 2       & 81.26          \\
    I-ViT-S/16 & Prompt 1       & \textbf{81.52}   \\
    \bottomrule
    \end{tabular}
    \label{tab:comparsion on prompts}
\end{table}
The impact of different prompts on VFM performance is compared in Table \ref{tab:comparsion on prompts}. Generally, incorporating concepts from VLMs enhances performance. However, the extent of improvement correlates with the prompts' ability to focus on the object. A stronger focus on the object indicates a more precise cognitive process, leading to superior outcomes.


For dense prediction tasks requiring multi-objective awareness, we synergistically combine \textbf{Language Prompts} and \textbf{Visual Prompts} to guide the VLM's focus. The language prompt is structured as: 
\begin{quote}
\textit{``If you are doing an object detection task, please tell me if there is/are \{tgt\_obj\} in this image. The answer format is as follows: Yes, there is/are \{tgt\_obj\} in the image, and the background is \{background\}.''}
\end{quote}
where \{tgt\_obj\} and \{background\} are dynamically replaced with target objects (e.g., \textit{dog, car}) and contextual attributes (e.g., \textit{grass, urban}) from task-specific annotations. This interrogative template forces the VLM to explicitly verify each object's presence and environmental context.

\begin{figure}[!htpb]
\centering
\includegraphics[width=0.5\textwidth]{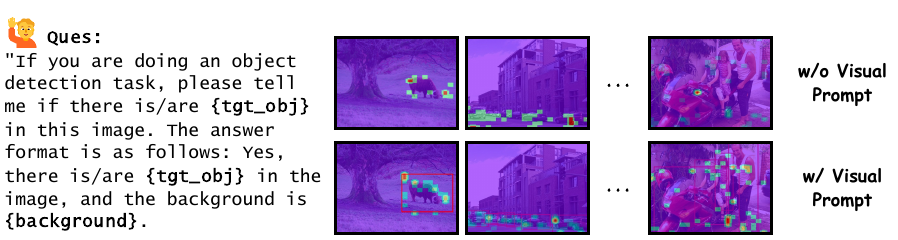}
\caption{
\textbf{Interaction Extraction for Dense Prediction.} The process uses $ \left\{ \textbf{tgt\_obj} \right\} $ to aggregate concepts.}
\label{fig:concept_extraction}
\end{figure}

Concurrently, \textbf{Visual Prompts} are implemented by overlaying ground-truth bounding boxes on input images (Figure~\ref{fig:concept_extraction}), spatially constraining the VLM's attention to instance regions. These boxes act as positional anchors during cross-modal interaction computation, reducing distraction from cluttered backgrounds.

As evidenced by Table~\ref{tab:prompt_ablation}, integrating visual prompts yields gains: +2.0 mAP, establishing a closed-loop feedback between linguistic verification and visual grounding.

\begin{table}[t]
\centering
\caption{Impact of Visual Prompts on COCO val2017 \label{tab:prompt_ablation}}
\begin{tabular}{lc}
\toprule
\textbf{Configuration} & \textbf{mAP} \\
\midrule
Language Prompt Only & 41.6\\
Language + Visual Prompts & 43.6 (+2.0) \\
\bottomrule
\end{tabular}
\vspace{-6pt}
\end{table}

\section{Human evaluation}
\label{sec:hum_eval}

\begin{figure}[!tpb]
\centering
\includegraphics[width=1\linewidth]{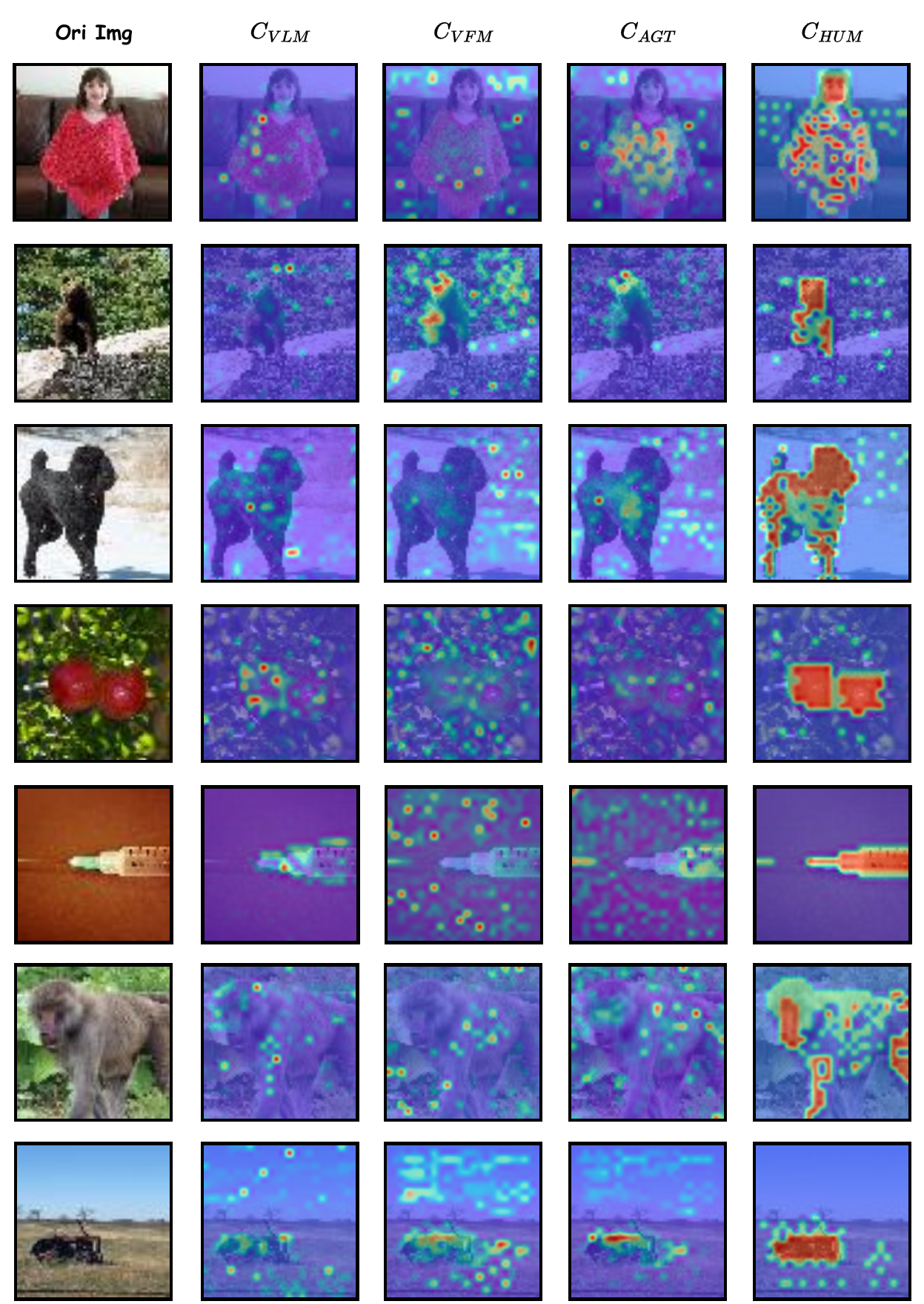}
\caption{
\textbf{Examples of Human Evaluation.} The figure illustrates the annotation process, where participants labeled image tokens based on their direct or indirect influence on the image label.}

\label{fig:human_evaluation_examples}
\end{figure}

To ensure the reliability and objectivity of our evaluation, we employed a double-blind assessment methodology. All participants were senior researchers with recognized expertise in the field. During the annotation process, participants were blinded to the annotations of their peers as well as to the results of $ C_{VLM} $, $ C_{AGT} $, and $ C_{VFM} $. However, they were provided with the corresponding image labels to maintain objectivity. Participants were tasked with annotating two categories: (1) image tokens that directly determine the label with high confidence (1.0), and (2) tokens that indirectly influence the label with low confidence (0.5), such as background regions. 20 participants provided $\sim1k$ annotated results, which were used to assess the similarity between different concepts and human annotations. Examples of the evaluation process are illustrated in Figure \ref{fig:human_evaluation_examples}, and the findings align with those presented in Section \ref{sec:vis_concept}.

\section{Concept weights on different layers}

\begin{figure}[!htpb]
\centering
\includegraphics[width=0.9\linewidth]{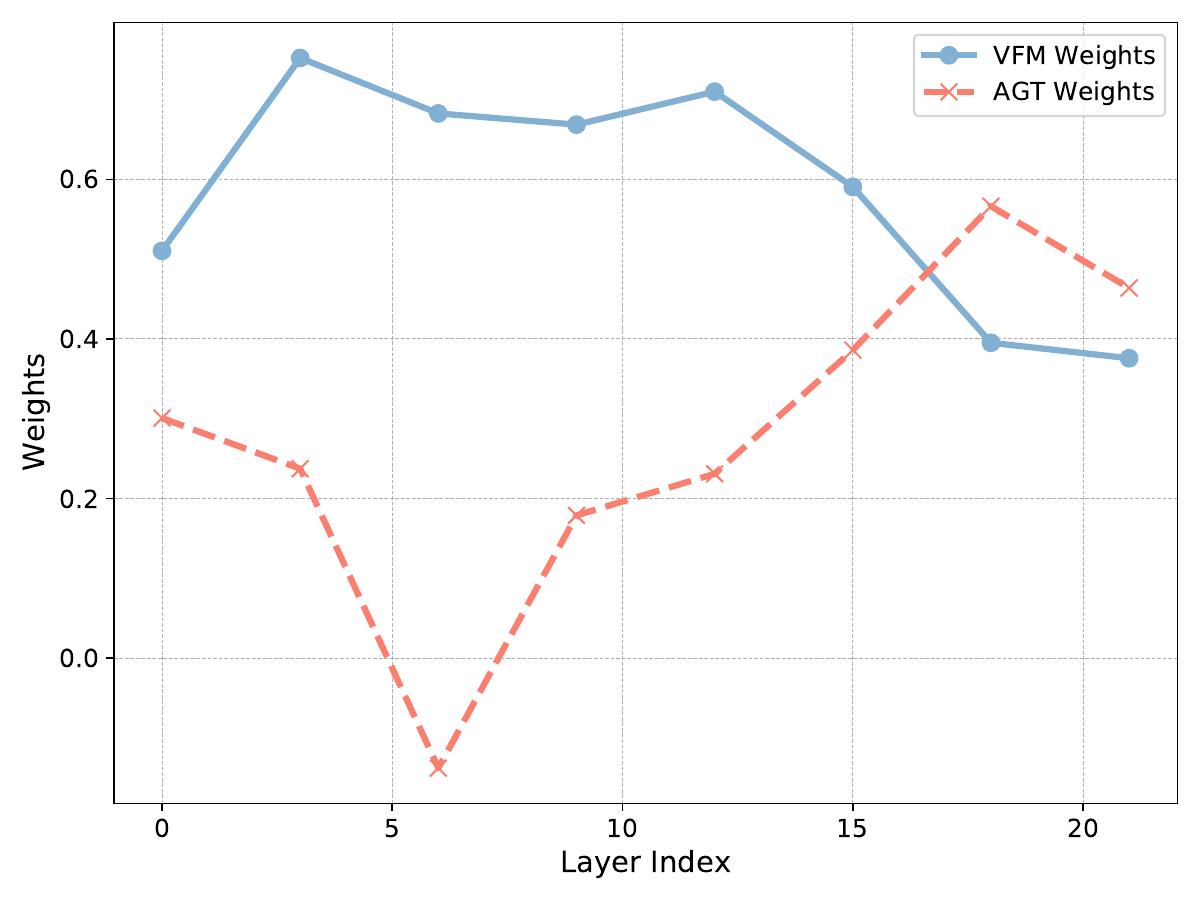}
\caption{
\textbf{Weights of Different Concepts Across Layers.} The model rapidly captures task-related concepts in early layers and further refines them in later layers using concepts from the VLM.}
\label{fig:trend}
\end{figure}

Although we incorporate additional $ C_{VLM} $ supervision into every layer of the Transformer in I-ViT, the model’s preference for $ C_{VFM} $ and $ C_{VLM} $ evolves with increasing model depth, as shown in Figure \ref{fig:trend}. We posit that the primary reason for this phenomenon is the rapid acquisition of task-related concepts in the initial layers, followed by their refinement in the deeper layers through the integration of VLM concepts. This observation suggests the following:

\begin{enumerate}[label=\arabic*.]
    \item The Gated Control Network (GCN) effectively modulates the weights between different concepts, preventing over-reliance on any single interaction and ensuring balanced consideration.
    \item Different concepts are complementary rather than interchangeable.
\end{enumerate}

By effectively leveraging concepts from various sources, the model achieves improved generalization and robustness.

%% file: main.bbl
\begin{thebibliography}{54}
\providecommand{\natexlab}[1]{#1}
\providecommand{\url}[1]{\texttt{#1}}
\expandafter\ifx\csname urlstyle\endcsname\relax
  \providecommand{\doi}[1]{doi: #1}\else
  \providecommand{\doi}{doi: \begingroup \urlstyle{rm}\Url}\fi

\bibitem[huh()]{huh2024platonic}
The platonic representation hypothesis.
\newblock \emph{arXiv preprint arXiv:2405.07987}.

\bibitem[pan()]{pang2023frozen_reb}
Frozen transformers in language models are effective visual encoder layers.
\newblock \emph{arXiv preprint arXiv:2310.12973}.

\bibitem[she()]{shen2023can}
Can the inference logic of large language models be disentangled into symbolic concepts?
\newblock \emph{arXiv preprint arXiv:2304.01083}.

\bibitem[Abnar and Zuidema(2022)]{abnar2022quantifying}
S Abnar and W Zuidema.
\newblock Quantifying attention flow in transformers. arxiv 2020.
\newblock \emph{arXiv preprint arXiv:2005.00928}, 2022.

\bibitem[Achiam et~al.(2023)Achiam, Adler, Agarwal, Ahmad, Akkaya, Aleman, Almeida, Altenschmidt, Altman, Anadkat, et~al.]{achiam2023gpt}
Josh Achiam, Steven Adler, Sandhini Agarwal, Lama Ahmad, Ilge Akkaya, Florencia~Leoni Aleman, Diogo Almeida, Janko Altenschmidt, Sam Altman, Shyamal Anadkat, et~al.
\newblock Gpt-4 technical report.
\newblock \emph{arXiv preprint arXiv:2303.08774}, 2023.

\bibitem[Anil et~al.(2023)Anil, Borgeaud, Wu, Alayrac, Yu, Soricut, Schalkwyk, Dai, Hauth, Millican, et~al.]{anil2023gemini}
Rohan Anil, Sebastian Borgeaud, Yonghui Wu, Jean-Baptiste Alayrac, Jiahui Yu, Radu Soricut, Johan Schalkwyk, Andrew~M Dai, Anja Hauth, Katie Millican, et~al.
\newblock Gemini: A family of highly capable multimodal models.
\newblock \emph{arXiv preprint arXiv:2312.11805}, 1, 2023.

\bibitem[Assran et~al.(2023)Assran, Duval, Misra, Bojanowski, Vincent, Rabbat, LeCun, and Ballas]{assran2023self}
Mahmoud Assran, Quentin Duval, Ishan Misra, Piotr Bojanowski, Pascal Vincent, Michael Rabbat, Yann LeCun, and Nicolas Ballas.
\newblock Self-supervised learning from images with a joint-embedding predictive architecture.
\newblock In \emph{Proceedings of the IEEE/CVF Conference on Computer Vision and Pattern Recognition}, pages 15619--15629, 2023.

\bibitem[Bangalath et~al.(2022)Bangalath, Maaz, Khattak, Khan, and Shahbaz~Khan]{bangalath2022bridging}
Hanoona Bangalath, Muhammad Maaz, Muhammad~Uzair Khattak, Salman~H Khan, and Fahad Shahbaz~Khan.
\newblock Bridging the gap between object and image-level representations for open-vocabulary detection.
\newblock \emph{Advances in Neural Information Processing Systems}, 35:\penalty0 33781--33794, 2022.

\bibitem[Cai et~al.(2024)Cai, Zhang, He, He, Tong, Gan, Wang, and Bai]{cai2024llava}
Yuxuan Cai, Jiangning Zhang, Haoyang He, Xinwei He, Ao Tong, Zhenye Gan, Chengjie Wang, and Xiang Bai.
\newblock Llava-kd: A framework of distilling multimodal large language models.
\newblock \emph{arXiv preprint arXiv:2410.16236}, 2024.

\bibitem[Carion et~al.(2020)Carion, Massa, Synnaeve, Usunier, Kirillov, and Zagoruyko]{detr}
Nicolas Carion, Francisco Massa, Gabriel Synnaeve, Nicolas Usunier, Alexander Kirillov, and Sergey Zagoruyko.
\newblock End-to-end object detection with transformers.
\newblock In \emph{European conference on computer vision}, pages 213--229. Springer, 2020.

\bibitem[Chefer et~al.(2021{\natexlab{a}})Chefer, Gur, and Wolf]{chefer2021generic}
Hila Chefer, Shir Gur, and Lior Wolf.
\newblock Generic attention-model explainability for interpreting bi-modal and encoder-decoder transformers.
\newblock In \emph{Proceedings of the IEEE/CVF International Conference on Computer Vision}, pages 397--406, 2021{\natexlab{a}}.

\bibitem[Chefer et~al.(2021{\natexlab{b}})Chefer, Gur, and Wolf]{chefer2021transformer}
Hila Chefer, Shir Gur, and Lior Wolf.
\newblock Transformer interpretability beyond attention visualization.
\newblock In \emph{Proceedings of the IEEE/CVF conference on computer vision and pattern recognition}, pages 782--791, 2021{\natexlab{b}}.

\bibitem[Chen et~al.(2023)Chen, Yang, Vondrick, and Mao]{chen2023interpreting}
Haozhe Chen, Junfeng Yang, Carl Vondrick, and Chengzhi Mao.
\newblock Interpreting and controlling vision foundation models via text explanations.
\newblock \emph{arXiv preprint arXiv:2310.10591}, 2023.

\bibitem[Chen et~al.(2021)Chen, Liu, Zhao, and Jia]{chen2021distilling}
Pengguang Chen, Shu Liu, Hengshuang Zhao, and Jiaya Jia.
\newblock Distilling knowledge via knowledge review.
\newblock In \emph{Proceedings of the IEEE/CVF conference on computer vision and pattern recognition}, pages 5008--5017, 2021.

\bibitem[Chen et~al.(2024)Chen, Wu, Wang, Su, Chen, Xing, Zhong, Zhang, Zhu, Lu, et~al.]{chen2024internvl}
Zhe Chen, Jiannan Wu, Wenhai Wang, Weijie Su, Guo Chen, Sen Xing, Muyan Zhong, Qinglong Zhang, Xizhou Zhu, Lewei Lu, et~al.
\newblock Internvl: Scaling up vision foundation models and aligning for generic visual-linguistic tasks.
\newblock In \emph{Proceedings of the IEEE/CVF Conference on Computer Vision and Pattern Recognition}, pages 24185--24198, 2024.

\bibitem[Dai et~al.(2023)Dai, Li, Li, Tiong, Zhao, Wang, Li, Fung, and Hoi]{dai2023instructblip}
Wenliang Dai, Junnan Li, Dongxu Li, Anthony Tiong, Junqi Zhao, Weisheng Wang, Boyang Li, Pascale Fung, and Steven Hoi.
\newblock Instruct{BLIP}: Towards general-purpose vision-language models with instruction tuning.
\newblock In \emph{Thirty-seventh Conference on Neural Information Processing Systems}, 2023.

\bibitem[Dosovitskiy(2020)]{dosovitskiy2020image}
Alexey Dosovitskiy.
\newblock An image is worth 16x16 words: Transformers for image recognition at scale.
\newblock \emph{arXiv preprint arXiv:2010.11929}, 2020.

\bibitem[Gao et~al.(2024)Gao, Geng, Zhang, Ma, Fang, Zhang, Li, and Qiao]{gao2024clip}
Peng Gao, Shijie Geng, Renrui Zhang, Teli Ma, Rongyao Fang, Yongfeng Zhang, Hongsheng Li, and Yu Qiao.
\newblock Clip-adapter: Better vision-language models with feature adapters.
\newblock \emph{International Journal of Computer Vision}, 132\penalty0 (2):\penalty0 581--595, 2024.

\bibitem[Gu et~al.(2021)Gu, Lin, Kuo, and Cui]{gu2021open}
Xiuye Gu, Tsung-Yi Lin, Weicheng Kuo, and Yin Cui.
\newblock Open-vocabulary object detection via vision and language knowledge distillation.
\newblock \emph{arXiv preprint arXiv:2104.13921}, 2021.

\bibitem[Hinton(2015)]{hinton2015distilling}
Geoffrey Hinton.
\newblock Distilling the knowledge in a neural network.
\newblock \emph{arXiv preprint arXiv:1503.02531}, 2015.

\bibitem[Hu et~al.(2024)Hu, Gu, Yu, Yu, Li, You, Lu, and Dong]{hu2024interpreting}
Jinfan Hu, Jinjin Gu, Shiyao Yu, Fanghua Yu, Zheyuan Li, Zhiyuan You, Chaochao Lu, and Chao Dong.
\newblock Interpreting low-level vision models with causal effect maps.
\newblock \emph{arXiv preprint arXiv:2407.19789}, 2024.

\bibitem[Kim and Chae(2024)]{kim2024does}
Seonggyeom Kim and Dong-Kyu Chae.
\newblock What does a model really look at?: Extracting model-oriented concepts for explaining deep neural networks.
\newblock \emph{IEEE Transactions on Pattern Analysis and Machine Intelligence}, 2024.

\bibitem[Lake and Baroni(2023)]{lake2023human}
Brenden~M Lake and Marco Baroni.
\newblock Human-like systematic generalization through a meta-learning neural network.
\newblock \emph{Nature}, 623\penalty0 (7985):\penalty0 115--121, 2023.

\bibitem[Li et~al.(2022)Li, Zhang, Liu, Guo, Ni, and Zhang]{li2022dn}
Feng Li, Hao Zhang, Shilong Liu, Jian Guo, Lionel~M Ni, and Lei Zhang.
\newblock Dn-detr: Accelerate detr training by introducing query denoising.
\newblock In \emph{Proceedings of the IEEE/CVF conference on computer vision and pattern recognition}, pages 13619--13627, 2022.

\bibitem[Li et~al.(2023)Li, Li, Savarese, and Hoi]{li2023blip}
Junnan Li, Dongxu Li, Silvio Savarese, and Steven Hoi.
\newblock Blip-2: Bootstrapping language-image pre-training with frozen image encoders and large language models.
\newblock In \emph{International conference on machine learning}, pages 19730--19742. PMLR, 2023.

\bibitem[Li and Zhang(2023)]{li2023does}
Mingjie Li and Quanshi Zhang.
\newblock Does a neural network really encode symbolic concepts?
\newblock In \emph{International conference on machine learning}, pages 20452--20469. PMLR, 2023.

\bibitem[Li et~al.(2024)Li, Zhang, Wang, Zhong, Chen, Chu, Liu, and Jia]{li2024mini}
Yanwei Li, Yuechen Zhang, Chengyao Wang, Zhisheng Zhong, Yixin Chen, Ruihang Chu, Shaoteng Liu, and Jiaya Jia.
\newblock Mini-gemini: Mining the potential of multi-modality vision language models.
\newblock \emph{arXiv preprint arXiv:2403.18814}, 2024.

\bibitem[Lin et~al.(2024)Lin, Chen, Zhao, and Wang]{lin2024advancing}
Mengying Lin, Yaran Chen, Dongbin Zhao, and Zhaoran Wang.
\newblock Advancing object goal navigation through llm-enhanced object affinities transfer.
\newblock \emph{arXiv preprint arXiv:2403.09971}, 2024.

\bibitem[Liu et~al.(2024{\natexlab{a}})Liu, Li, Li, and Lee]{liu2024improved}
Haotian Liu, Chunyuan Li, Yuheng Li, and Yong~Jae Lee.
\newblock Improved baselines with visual instruction tuning.
\newblock In \emph{Proceedings of the IEEE/CVF Conference on Computer Vision and Pattern Recognition}, pages 26296--26306, 2024{\natexlab{a}}.

\bibitem[Liu et~al.(2024{\natexlab{b}})Liu, Li, Wu, and Lee]{liu2024visual}
Haotian Liu, Chunyuan Li, Qingyang Wu, and Yong~Jae Lee.
\newblock Visual instruction tuning.
\newblock \emph{Advances in neural information processing systems}, 36, 2024{\natexlab{b}}.

\bibitem[Liu et~al.(2022)Liu, Li, Zhang, Yang, Qi, Su, Zhu, and Zhang]{liu2022dab}
Shilong Liu, Feng Li, Hao Zhang, Xiao Yang, Xianbiao Qi, Hang Su, Jun Zhu, and Lei Zhang.
\newblock Dab-detr: Dynamic anchor boxes are better queries for detr.
\newblock \emph{arXiv preprint arXiv:2201.12329}, 2022.

\bibitem[Liu et~al.(2021)Liu, Lin, Cao, Hu, Wei, Zhang, Lin, and Guo]{liu2021swin}
Ze Liu, Yutong Lin, Yue Cao, Han Hu, Yixuan Wei, Zheng Zhang, Stephen Lin, and Baining Guo.
\newblock Swin transformer: Hierarchical vision transformer using shifted windows.
\newblock In \emph{Proceedings of the IEEE/CVF international conference on computer vision}, pages 10012--10022, 2021.

\bibitem[Lyu et~al.(2022)Lyu, Liang, Deng, Salakhutdinov, and Morency]{lyu2022dime}
Yiwei Lyu, Paul~Pu Liang, Zihao Deng, Ruslan Salakhutdinov, and Louis-Philippe Morency.
\newblock Dime: Fine-grained interpretations of multimodal models via disentangled local explanations.
\newblock In \emph{Proceedings of the 2022 AAAI/ACM Conference on AI, Ethics, and Society}, pages 455--467, 2022.

\bibitem[Ma et~al.(2023)Ma, Bai, Zhong, Zhang, Yao, and Mei]{ma2023visualizing}
Jie Ma, Yalong Bai, Bineng Zhong, Wei Zhang, Ting Yao, and Tao Mei.
\newblock Visualizing and understanding patch interactions in vision transformer.
\newblock \emph{IEEE Transactions on Neural Networks and Learning Systems}, 2023.

\bibitem[Ma et~al.(2022)Ma, Luo, Gao, Li, Chen, Wang, Zhang, and Hu]{ma2022open}
Zongyang Ma, Guan Luo, Jin Gao, Liang Li, Yuxin Chen, Shaoru Wang, Congxuan Zhang, and Weiming Hu.
\newblock Open-vocabulary one-stage detection with hierarchical visual-language knowledge distillation.
\newblock In \emph{Proceedings of the IEEE/CVF Conference on Computer Vision and Pattern Recognition}, pages 14074--14083, 2022.

\bibitem[Meng et~al.(2021{\natexlab{a}})Meng, Chen, Fan, Zeng, Li, Yuan, Sun, and Wang]{conditional_detr}
Depu Meng, Xiaokang Chen, Zejia Fan, Gang Zeng, Houqiang Li, Yuhui Yuan, Lei Sun, and Jingdong Wang.
\newblock Conditional detr for fast training convergence.
\newblock In \emph{Proceedings of the IEEE/CVF International Conference on Computer Vision (ICCV)}, pages 3651--3660, 2021{\natexlab{a}}.

\bibitem[Meng et~al.(2021{\natexlab{b}})Meng, Chen, Fan, Zeng, Li, Yuan, Sun, and Wang]{meng2021conditional}
Depu Meng, Xiaokang Chen, Zejia Fan, Gang Zeng, Houqiang Li, Yuhui Yuan, Lei Sun, and Jingdong Wang.
\newblock Conditional detr for fast training convergence.
\newblock In \emph{Proceedings of the IEEE/CVF international conference on computer vision}, pages 3651--3660, 2021{\natexlab{b}}.

\bibitem[Niu et~al.(2024)Niu, Liu, Bi, Feng, Peng, and Chen]{niu2024large}
Qian Niu, Junyu Liu, Ziqian Bi, Pohsun Feng, Benji Peng, and Keyu Chen.
\newblock Large language models and cognitive science: A comprehensive review of similarities, differences, and challenges.
\newblock \emph{arXiv preprint arXiv:2409.02387}, 2024.

\bibitem[Pang et~al.(2023)Pang, Xie, Man, and Wang]{pang2023frozen}
Ziqi Pang, Ziyang Xie, Yunze Man, and Yu-Xiong Wang.
\newblock Frozen transformers in language models are effective visual encoder layers.
\newblock \emph{arXiv preprint arXiv:2310.12973}, 2023.

\bibitem[Rasekh et~al.(2024)Rasekh, Ranjbar, Heidari, and Nejdl]{rasekh2024ecor}
Ali Rasekh, Sepehr~Kazemi Ranjbar, Milad Heidari, and Wolfgang Nejdl.
\newblock Ecor: Explainable clip for object recognition.
\newblock \emph{arXiv preprint arXiv:2404.12839}, 2024.

\bibitem[Shu et~al.(2024)Shu, Liao, Zhuo, Xu, Zhang, Shi, Chen, Zhong, He, Fu, et~al.]{shu2024llava}
Fangxun Shu, Yue Liao, Le Zhuo, Chenning Xu, Guanghao Zhang, Haonan Shi, Long Chen, Tao Zhong, Wanggui He, Siming Fu, et~al.
\newblock Llava-mod: Making llava tiny via moe knowledge distillation.
\newblock \emph{arXiv preprint arXiv:2408.15881}, 2024.

\bibitem[Stan et~al.(2024)Stan, Rohekar, Gurwicz, Olson, Bhiwandiwalla, Aflalo, Wu, Duan, Tseng, and Lal]{stan2024lvlm}
Gabriela Ben~Melech Stan, Raanan~Yehezkel Rohekar, Yaniv Gurwicz, Matthew~Lyle Olson, Anahita Bhiwandiwalla, Estelle Aflalo, Chenfei Wu, Nan Duan, Shao-Yen Tseng, and Vasudev Lal.
\newblock Lvlm-intrepret: An interpretability tool for large vision-language models.
\newblock \emph{arXiv preprint arXiv:2404.03118}, 2024.

\bibitem[Vaswani(2017)]{vaswani2017attention}
A Vaswani.
\newblock Attention is all you need.
\newblock \emph{Advances in Neural Information Processing Systems}, 2017.

\bibitem[Wang et~al.(2023)Wang, Lv, Yu, Hong, Qi, Wang, Ji, Yang, Zhao, Song, et~al.]{wang2023cogvlm}
Weihan Wang, Qingsong Lv, Wenmeng Yu, Wenyi Hong, Ji Qi, Yan Wang, Junhui Ji, Zhuoyi Yang, Lei Zhao, Xixuan Song, et~al.
\newblock Cogvlm: Visual expert for pretrained language models.
\newblock \emph{arXiv preprint arXiv:2311.03079}, 2023.

\bibitem[Xiao et~al.(2024)Xiao, Wu, Wang, Li, Zhou, and Guo]{xiao2024seeing}
Xin Xiao, Bohong Wu, Jiacong Wang, Chunyuan Li, Xun Zhou, and Haoyuan Guo.
\newblock Seeing the image: Prioritizing visual correlation by contrastive alignment.
\newblock \emph{arXiv preprint arXiv:2405.17871}, 2024.

\bibitem[Yao et~al.(2024)Yao, Li, Ren, Wang, Liu, Sun, and Hou]{yao2024deco}
Linli Yao, Lei Li, Shuhuai Ren, Lean Wang, Yuanxin Liu, Xu Sun, and Lu Hou.
\newblock Deco: Decoupling token compression from semantic abstraction in multimodal large language models.
\newblock \emph{arXiv preprint arXiv:2405.20985}, 2024.

\bibitem[Zagoruyko and Komodakis(2016)]{zagoruyko2016paying}
Sergey Zagoruyko and Nikos Komodakis.
\newblock Paying more attention to attention: Improving the performance of convolutional neural networks via attention transfer.
\newblock \emph{arXiv preprint arXiv:1612.03928}, 2016.

\bibitem[Zhang et~al.(2022)Zhang, Li, Liu, Zhang, Su, Zhu, Ni, and Shum]{zhang2022dino}
Hao Zhang, Feng Li, Shilong Liu, Lei Zhang, Hang Su, Jun Zhu, Lionel~M Ni, and Heung-Yeung Shum.
\newblock Dino: Detr with improved denoising anchor boxes for end-to-end object detection.
\newblock \emph{arXiv preprint arXiv:2203.03605}, 2022.

\bibitem[Zhang et~al.(2021)Zhang, Fang, Zhang, Gao, Li, Dai, Qiao, and Li]{zhang2021tip}
Renrui Zhang, Rongyao Fang, Wei Zhang, Peng Gao, Kunchang Li, Jifeng Dai, Yu Qiao, and Hongsheng Li.
\newblock Tip-adapter: Training-free clip-adapter for better vision-language modeling.
\newblock \emph{arXiv preprint arXiv:2111.03930}, 2021.

\bibitem[Zhang et~al.(2024)Zhang, Fan, Ma, Zheng, Huang, Cheng, Gudovskiy, Okuno, Nakata, Keutzer, et~al.]{zhang2024sparsevlm}
Yuan Zhang, Chun-Kai Fan, Junpeng Ma, Wenzhao Zheng, Tao Huang, Kuan Cheng, Denis Gudovskiy, Tomoyuki Okuno, Yohei Nakata, Kurt Keutzer, et~al.
\newblock Sparsevlm: Visual token sparsification for efficient vision-language model inference.
\newblock \emph{arXiv preprint arXiv:2410.04417}, 2024.

\bibitem[Zhou et~al.(2022{\natexlab{a}})Zhou, Yang, Loy, and Liu]{zhou2022conditional}
Kaiyang Zhou, Jingkang Yang, Chen~Change Loy, and Ziwei Liu.
\newblock Conditional prompt learning for vision-language models.
\newblock In \emph{Proceedings of the IEEE/CVF conference on computer vision and pattern recognition}, pages 16816--16825, 2022{\natexlab{a}}.

\bibitem[Zhou et~al.(2022{\natexlab{b}})Zhou, Yang, Loy, and Liu]{zhou2022learning}
Kaiyang Zhou, Jingkang Yang, Chen~Change Loy, and Ziwei Liu.
\newblock Learning to prompt for vision-language models.
\newblock \emph{International Journal of Computer Vision}, 130\penalty0 (9):\penalty0 2337--2348, 2022{\natexlab{b}}.

\bibitem[Zhu et~al.(2023)Zhu, Chen, Shen, Li, and Elhoseiny]{zhu2023minigpt}
Deyao Zhu, Jun Chen, Xiaoqian Shen, Xiang Li, and Mohamed Elhoseiny.
\newblock Minigpt-4: Enhancing vision-language understanding with advanced large language models.
\newblock \emph{arXiv preprint arXiv:2304.10592}, 2023.

\bibitem[Zhu et~al.(2020)Zhu, Su, Lu, Li, Wang, and Dai]{zhu2020deformable}
Xizhou Zhu, Weijie Su, Lewei Lu, Bin Li, Xiaogang Wang, and Jifeng Dai.
\newblock Deformable detr: Deformable transformers for end-to-end object detection.
\newblock \emph{arXiv preprint arXiv:2010.04159}, 2020.

\end{thebibliography}
